\documentclass{article}
\usepackage[final]{colm2026_conference}

\usepackage{microtype}
\usepackage{hyperref}
\usepackage{url}
\usepackage{booktabs}
\usepackage{graphicx}
\usepackage{multirow}
\usepackage{fvextra}
\usepackage{float}
\usepackage{amsmath}
\usepackage{amssymb}
\usepackage{lineno}

\definecolor{darkblue}{rgb}{0, 0, 0.5}
\hypersetup{colorlinks=true, citecolor=darkblue, linkcolor=darkblue, urlcolor=darkblue}

\title{Claim-Level Reliability Assessment for Efficient Test-Time Reasoning\thanks{Code is available at \url{https://github.com/WeiboAI/CLR}.}}

\author{\textbf{Sen Xu\quad Wei Wang\quad Shixi Liu\quad Jixin Min\quad Yingwei Dai}\\
\textbf{Zhibin Yin\quad Yirong Chen\quad Junlin Zhang}\\
Sina Weibo Inc.\\
\texttt{\{xusen1,junlin6\}@staff.weibo.com}
}

\begin{document}

\ifcolmsubmission
\linenumbers
\fi

\maketitle

\begin{abstract}
We propose claim-level falsification as a principle for test-time scaling and instantiate it through Claim-Level Reliability Assessment (CLR), a training-free framework that reallocates test-time compute from additional solution sampling to targeted verification. Since whole-trace evaluation often obscures decisive errors due to signal dilution from routine tokens, CLR condenses each reasoning trace into a compact set of decision-critical claims, thereby isolating its logical anchors. Furthermore, recognizing the inherent difficulty of generating entirely correct solutions under fixed model capabilities, CLR shifts the focus to semantic falsification. This approach exploits a fundamental asymmetry between solution construction and claim refutation. Constructing a valid solution requires a flawless reasoning path, whereas refuting an incorrect claim requires identifying only a single decisive flaw. This targeted search for negative evidence systematically compresses the survival space of high-confidence incorrect traces, effectively suppressing erroneous consensus via nonlinear reliability scoring. Across four LLMs and four reasoning benchmarks under matched budgets, CLR generally improves upon pass@1 and self-consistency. On GPT-OSS-20B/CMIMC25, for instance, CLR exceeds pass@1 by 27.15 percentage-points and raises self-consistency accuracy from 77.50\% to 82.19\% with 37.0\% fewer tokens. 
\end{abstract}

\section{Introduction}

Test-time scaling has emerged as an effective paradigm for enhancing the reasoning capabilities of large language models (LLMs) without parameter updates \citep{snell2025scaling,muennighoff2025s1,wang2022consK}. Existing methods can be broadly categorized based on where reliability signals are applied during the reasoning process. The first category derives intrinsic signals, such as token probabilities, entropy, or hidden states \citep{fu2025deepconf,wang2025latent}, to estimate the overall uncertainty of a single completed trace. The second operates across multiple parallel traces, producing final predictions via answer aggregation or Best-of-N selection \citep{cobbe2021training,toshniwal2025genselect,weng2023large,zhao2025sample,singhisolve,chensets}. The third intervenes during generation by using explicit evaluation signals to guide structured searches like branching, pruning, or backtracking \citep{yao2023tree}. Despite their differences, these approaches share a fundamental bottleneck. As inference budgets grow, extracting accurate and highly discriminative reliability signals remains challenging, which ultimately limits the extent to which additional compute translates into trustworthy answers.


Crucially, existing reliability signals often fail to capture the decision-critical semantic content that dictates answer correctness. Methods relying on token probabilities or internal states implicitly treat statistical confidence as a proxy for logical reliability. However, high statistical confidence does not guarantee logical soundness, and a highly confident trace may still harbor a decisive flaw \citep{xiong2024can}. Furthermore, whole-trace evaluation suffers from severe signal dilution. Because most tokens correspond to routine reasoning steps, they create a weakly discriminative background that degrades the signal-to-noise ratio. This background often obscures localized but fatal mistakes within an otherwise plausible trace. While step-level verification can isolate such errors, it is computationally exhaustive and typically requires process-level supervision or separately trained verifiers \citep{lightman2024let}. These limitations highlight the critical need for an intermediate granularity that bridges the gap between whole-trace and step-by-step evaluation by directly targeting the semantic content anchoring a reasoning trace.


Building on these observations, we formulate claim-level falsification as a principle for test-time scaling and instantiate it through Claim-Level Reliability Assessment (CLR), a training-free framework for consensus-based answer aggregation. CLR extracts more discriminative reliability signals through two complementary mechanisms: \textbf{\textit{(1) Claim-level evaluation.}} Instead of evaluating the full trace, CLR condenses it into a compact set of decision-critical claims. By filtering out routine tokens that dilute the reliability signal, this representation directs test-time compute exclusively toward the logical anchors that determine answer correctness. \textbf{\textit{(2) Falsification-based verification.}} Treating each claim as a falsifiable proposition, CLR prompts the model to actively search for disconfirming evidence. This establishes a fundamental asymmetry: while constructing a correct solution demands a completely valid reasoning path, refuting an incorrect claim requires identifying only a single decisive flaw. By replacing open-ended re-solving with targeted refutation, CLR effectively repurposes the base model into a rigorous verifier. Crucially, this focused, one-sided check can expose critical errors overlooked during generation, even within erroneous traces produced with high statistical confidence.

Operationally, CLR follows a two-stage inference pipeline. In the first stage, it samples \(K\) solution traces, each accompanied by a compact set of decision-critical claims. In the second stage, the same model independently verifies the claims associated with each trace using only the original problem and the extracted claims. We use Cons@\(K\) to denote self-consistency \citep{wang2022consK} with \(K\) sampled traces. CLR@\(K\) therefore requires \(K\) calls for solution generation and \(K\) calls for claim verification, matching the \(2K\) solution generation calls used by Cons@\(2K\). Finally, CLR maps claim-level verdicts into nonlinear trace-level reliability scores for reliability-weighted aggregation, allowing a reliable minority to overturn an incorrect consensus formed by a majority of flawed traces.

We evaluate CLR on four LLMs across multiple challenging reasoning benchmarks. Under matched model call budgets, CLR improves accuracy, token efficiency, or both across the evaluated model regimes. On HMMT25 \citep{hmmt2025}, CLR@32 raises the accuracy of Gemma-4-12B-it \citep{gemmateam2026gemma4technicalreport} from 76.67\% with Cons@64 to 88.75\%. On CMIMC25 \citep{cmimc2025}, CLR@32 improves GPT-OSS-20B \citep{agarwal2025gpt} from 77.50\% with Cons@64 to 82.19\% while using 37.0\% fewer generated tokens, outperforming Pass@1 by 27.15 percentage points.

\section{Method}
\label{sec:method}

\begin{figure}[t]
    \centering
    \includegraphics[width=\textwidth]{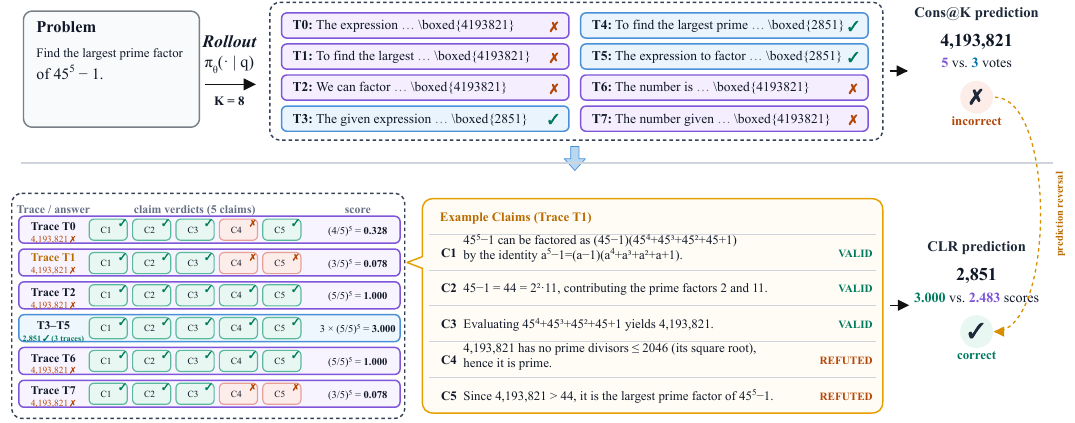}
    \caption{Overview of CLR through an illustrative example. Given the same
\(K=8\) sampled traces, count-based self-consistency selects the incorrect
answer \(4{,}193{,}821\) by a \(5\)-to-\(3\) majority. CLR instead applies
falsification-based checks to \(M=5\) decision-critical claims per trace and
uses the resulting claim-level outcomes to perform reliability-weighted
aggregation, recovering the correct answer \(2{,}851\).}
    \label{fig:clr_motivation}
\end{figure}

Given a problem \(q\) and a solution-sampling count \(K\), CLR independently samples
\(K\) traces under a fixed decoding configuration. For each trace, a Stage-1
request generates a complete reasoning trace \(t_k\), its final prediction \(y_k\),
and an ordered list of \(M\) decision-critical claims
\(C_k=(c_{k,1},\ldots,c_{k,M})\). A Stage-2 request then reuses the same model
to search for refutations using only \(q\) and \(C_k\). CLR converts the
resulting verdicts into a trace-level reliability score and aggregates
equivalent predictions by their total reliability. The second stage therefore
reweights sampled candidates without generating new predictions. Fig.~\ref{fig:clr_motivation} illustrates how claim-level reliability can
suppress an erroneous five-trace consensus and select a smaller, better-supported
answer group from the same eight samples.

\paragraph{Critical Claim Extraction.}
Stage 1 outputs \(y_k\) in a task-specific structured format and appends exactly
\(M\) concise intermediate claims whose failure would undermine the prediction,
excluding generic summaries and prediction restatements. Typical claims encode
intermediate conclusions, constraints, decision points, transformations, or
evidence linking \(q\) to \(y_k\), as illustrated by the factorization,
primality, and maximality statements in Fig.~\ref{fig:clr_motivation}. This
fixed-size representation compresses each trace into focused semantic
assertions. The full prompt appears in Appendix~\ref{app:stage1_prompt}.

\paragraph{Falsification-Based Claim Assessment.}
For each trace, CLR makes one joint Stage-2 request containing \(q\) and the
ordered claim list \(C_k\), without the original trace or final prediction as
separate inputs. The same model searches each claim for a decisive
contradiction, counterexample, factual or logical error, missing condition, or
unsupported inference, while also checking for conflicts across claims. We
encode the outcome as
\begin{equation}
v_{k,m} =
\begin{cases}
0, & \text{refuted, if a decisive flaw is found},\\
1, & \text{not refuted, otherwise}.
\end{cases}
\label{eq:claim_verdict}
\end{equation}
The \texttt{VALID} output token denotes only \emph{not refuted by this
assessment}, rather than formal proof of correctness. This one-sided objective
focuses compute on negative-evidence search, where one decisive witness can
refute a claim without constructing an alternative complete solution. We treat
this asymmetry as an inductive bias rather than a guarantee that falsification
is uniformly easier than generation. The complete prompt appears in
Appendix~\ref{app:stage2_prompt}.

\paragraph{Reliability Scoring and Aggregation.}
Let \(s_k\) be the fraction of claims that survive falsification. We map this
fraction to the trace score
\begin{equation}
r_k = s_k^M
    = \left( \frac{1}{M} \sum_{m=1}^{M} v_{k,m} \right)^M .
\label{eq:trace_reliability}
\end{equation}
For \(M>1\), the exponent makes a refuted decision-critical claim more
consequential than under linear averaging, increasingly suppressing a trace as
more of its decision-critical claims are refuted. Compared to linear averaging,
this nonlinear penalty reduces the aggregate influence of error-prone traces
even when they are numerically dominant, allowing a smaller but more reliable
answer group to overturn an incorrect majority, as illustrated in
Fig.~\ref{fig:clr_motivation}. This monotone transformation is a heuristic, not
a joint correctness probability, and does not assume claim independence. We
use the claim count \(M\) as the exponent in all experiments.

CLR extracts \(y_k\) with a task-specific output parser and omits traces without
a parsed prediction. It then partitions the predictions into equivalence groups
\(\mathcal{G}\) according to a task-appropriate equivalence criterion. For each
group \(G\in\mathcal{G}\), the reliability support is
\begin{equation}
R(G) = \sum_{k:y_k\in G} r_k,
\qquad
\widehat{y}=\operatorname{canon}\!\left(
    \underset{G\in\mathcal{G}}{\arg\max}\;R(G)
\right),
\label{eq:reliability_aggregation}
\end{equation}
where \(\operatorname{canon}(G)\) denotes the canonical candidate answer for
group \(G\). Eq.~\ref{eq:reliability_aggregation} reduces to ordinary
self-consistency when all traces receive the same positive score; otherwise,
refuted claims reduce a trace's influence on the final consensus. Ties,
including the all-zero-score case, are resolved by the earliest equivalence
group in sampling order.

\paragraph{Budget and Scope.}
One CLR@\(K\) flow uses \(K\) solution-generation requests and \(K\)
claim-assessment requests, so it matches the request count of
Cons@\(2K\). Request parity does not imply token parity because claim assessment
operates on \(q\) and \(M\) claims rather than generating another complete
solution; we therefore report both matched-request and realized-token comparisons.
CLR only reweights parsed candidates, so
\(\widehat{y}\in\{y_k:y_k\text{ is parsed}\}\) and cannot recover a correct
prediction absent from the Stage-1 samples. Its role is to convert candidate
coverage into a more reliable selection under a fixed request or token budget.

\section{Experiments}

\subsection{Experimental Setup}

\paragraph{Models and benchmarks.}
We evaluate CLR with Gemma-4-12B-it \citep{gemmateam2026gemma4technicalreport}, GPT-OSS-20B, GPT-OSS-120B \citep{agarwal2025gpt}, and
Qwen3.5-27B \citep{qwen2026qwen35} on multiple reasoning benchmarks:
HMMT25 \citep{hmmt2025}, HMMT26 \citep{hmmt2026}, CMIMC25 \citep{cmimc2025}, and Apex-shortlist \citep{dekoninck2026matharena&apexshortlist}.

\paragraph{Baselines and test-time settings.}
Our primary baseline is self-consistency \citep{wang2022consK}, which samples
\(K\) independent solutions and aggregates their final answers. We evaluate
Cons@\(K\) with \(K\in\{8,16,32,64\}\) and CLR@\(K\) with
\(K\in\{4,8,16,32\}\). Following the request accounting defined in
Section~\ref{sec:method}, our primary matched-request comparison is CLR@32
versus Cons@64.
Regular sampling uses a minimal solution prompt consisting of the problem and
a single step-by-step instruction. CLR Stage 1 extends this base prompt by
appending the claim-generation instructions; both templates are provided in
Appendix~\ref{app:prompts}.
Unless otherwise stated, CLR extracts \(M=5\) claims per trace, and its
accuracy is averaged over \(N=8\) independent complete flows.

\paragraph{Metrics.}
For regular sampling, single-rollout \(\mathrm{pass@1}\) is the accuracy
averaged over the 64 individual solution rollouts, whereas Cons@\(K\) reports
the accuracy obtained by aggregating \(K\) sampled solutions. For CLR,
CLR@\(K\) denotes a complete two-stage TTS flow comprising \(K\)
solution-generation calls and \(K\) claim-assessment calls. We report
\(\mathrm{pass@1}\) as the mean accuracy across \(N=8\) independent executions
of the full flow. We measure
efficiency using the average number of generated tokens per problem. CLR token
counts include both solution generation and claim-level assessment. To account
for their different output lengths, we report matched-request results together
with accuracy as a function of realized token consumption. Model-specific
decoding configurations are summarized in
Appendix~\ref{app:experimental_details}.

\subsection{Main Results}

\begin{table}[t]
\centering
\scriptsize
\setlength{\tabcolsep}{3.8pt}
\renewcommand{\arraystretch}{1.08}
\resizebox{\textwidth}{!}{%
\begin{tabular}{llc cc cc cc}
\toprule
\multirow{2}{*}{Model} & \multirow{2}{*}{Benchmark}
& \multirow{2}{*}{Pass@1}
& \multicolumn{2}{c}{Cons@64}
& \multicolumn{2}{c}{CLR@32}
& \multicolumn{2}{c}{$\Delta$ vs.\ Cons@64} \\
\cmidrule(lr){4-5} \cmidrule(lr){6-7} \cmidrule(lr){8-9}
& & & Acc. & Tok. ($\times 10^3$) & Acc. & Tok. ($\times 10^3$) & Acc. (pp) & Tok. (\%) \\
\midrule

\multirow{4}{*}{Gemma-4-12B-it}
& HMMT25 & 64.58 & 76.67 & 1342.3 & \textbf{88.75} & 1791.0 & \textbf{+12.08} & +33.4 \\
& HMMT26 & 57.72 & 69.70 & 1543.5 & \textbf{77.27} & 2203.8 & \textbf{+7.57} & +42.8 \\
& CMIMC25 & 55.23 & 68.75 & 1521.3 & \textbf{80.62} & 1858.9 & \textbf{+11.87} & +22.2 \\
& Apex-shortlist & 21.35 & 32.98 & 1829.4 & \textbf{40.10} & 2704.2 & \textbf{+7.12} & +47.8 \\
\midrule

\multirow{4}{*}{GPT-OSS-20B}
& HMMT25 & 58.23 & 80.00 & 968.1 & 79.58 & \textbf{616.4} & -0.42 & \textbf{-36.3} \\
& HMMT26 & 57.72 & 72.73 & 1037.1 & \textbf{73.48} & \textbf{624.5} & \textbf{+0.75} & \textbf{-39.8} \\
& CMIMC25 & 55.04 & 77.50 & 926.3 & \textbf{82.19} & \textbf{583.3} & \textbf{+4.69} & \textbf{-37.0} \\
& Apex-shortlist & 15.82 & 20.83 & 1292.4 & \textbf{24.22} & \textbf{795.0} & \textbf{+3.39} & \textbf{-38.5} \\
\midrule

\multirow{4}{*}{Qwen3.5-27B}
& HMMT25 & 91.41 & 93.33 & 2012.9 & \textbf{95.00} & \textbf{2004.8} & \textbf{+1.67} & \textbf{-0.4} \\
& HMMT26 & 82.53 & 90.91 & 2134.8 & 90.91 & \textbf{2083.3} & 0.00 & \textbf{-2.4} \\
& CMIMC25 & 85.90 & 95.00 & 2150.0 & \textbf{97.50} & \textbf{2096.9} & \textbf{+2.50} & \textbf{-2.5} \\
& Apex-shortlist & 54.72 & 72.92 & 3625.1 & \textbf{75.52} & \textbf{3099.3} & \textbf{+2.60} & \textbf{-14.5} \\
\midrule

\multirow{4}{*}{GPT-OSS-120B}
& HMMT25 & 61.46 & 86.67 & 459.1 & \textbf{87.08} & \textbf{350.1} & \textbf{+0.41} & \textbf{-23.7} \\
& HMMT26 & 59.19 & 72.73 & 457.3 & \textbf{75.38} & \textbf{350.7} & \textbf{+2.65} & \textbf{-23.3} \\
& CMIMC25 & 59.18 & 75.00 & 423.6 & \textbf{80.00} & \textbf{332.1} & \textbf{+5.00} & \textbf{-21.6} \\
& Apex-shortlist & 18.39 & 23.96 & 489.7 & \textbf{27.86} & \textbf{382.4} & \textbf{+3.90} & \textbf{-21.9} \\
\bottomrule
\end{tabular}%
}
\caption{Token--accuracy trade-off across four models and four reasoning
benchmarks under matched model-call budgets. CLR@32 and Cons@64
both use 64 model calls.}
\label{tab:main_results}
\end{table}

At the primary operating point reported in Tab.~\ref{tab:main_results},
accuracy deltas are given in
percentage points, while token deltas are relative changes; both are measured
against Cons@64. The accuracy--efficiency profile varies across base models.
For Gemma-4-12B-it, CLR improves accuracy on all four
benchmarks by 7.12--12.08 points, including an increase from
76.67\% to 88.75\% on HMMT25. These gains, however, come with 22.2--47.8\%
more generated tokens.
The results for GPT-OSS-20B follow a different pattern. CLR uses
36.3--39.8\% fewer tokens on every benchmark while improving accuracy in three
of four cases. On CMIMC25, it raises accuracy from 77.50\% to 82.19\% with
37.0\% fewer tokens. GPT-OSS-120B similarly improves accuracy by up to 5.00
percentage points while reducing token consumption by 21.6--23.7\%. For
Qwen3.5-27B, whose Cons@64 accuracy already exceeds 90\% on three benchmarks,
the remaining headroom is smaller. CLR matches or improves accuracy by up to
2.60 percentage points, with its largest token reduction of 14.5\% occurring
on Apex-shortlist.

The matched-call comparison indicates that test-time compute need not be spent
exclusively on generating more solutions. Reallocating half of the calls to
claim-level falsification can either
improve accuracy over count-based consensus or retain competitive accuracy with
substantially fewer generated tokens, with the dominant benefit depending on the
base model's operating regime.

\subsection{Decomposing the Gains of CLR}

\begin{table}[t]
\centering
\small
\setlength{\tabcolsep}{4.5pt}
\renewcommand{\arraystretch}{1.08}
\begin{tabular}{l cc cc c c}
\toprule
\multirow{2}{*}{Benchmark}
& \multicolumn{2}{c}{Regular Sampling}
& \multicolumn{2}{c}{CLR Stage 1 Only}
& \multicolumn{1}{c}{Full CLR}
& \multicolumn{1}{c}{Stage-2 Gain} \\
\cmidrule(lr){2-3}
\cmidrule(lr){4-5}
& Pass@1 & Cons@64
& Pass@1 & Cons@32
& CLR@32 & $\Delta$ (pp) \\
\midrule
HMMT25         & 58.23 & 80.00 & 53.67 & 75.10 & 79.58 & $+4.48$ \\
HMMT26         & 57.72 & 72.73 & 54.88 & 66.47 & 73.48 & $+7.01$ \\
CMIMC25        & 55.04 & 77.50 & 54.39 & 77.19 & 82.19 & $+5.00$ \\
Apex-shortlist & 15.82 & 20.83 & 14.68 & 17.93 & 24.22 & $+6.29$ \\
\bottomrule
\end{tabular}
\caption{Decomposition of CLR on GPT-OSS-20B. Stage-2 gain denotes the
accuracy difference between full CLR@32 and unweighted Stage-1 Cons@32 on the
same candidates.}
\label{tab:clr_decomposition}
\end{table}

Tab.~\ref{tab:clr_decomposition} separates Stage-1 claim prompting from
falsification-based reliability weighting. Unweighted aggregation over the same
32 candidates isolates Stage 2, while 64-sample self-consistency provides the
matched-call reference in Tab.~\ref{tab:main_results} because full CLR uses 32
calls per stage. The claim prompt alone reduces single-rollout accuracy by
0.65--4.56 percentage points relative to regular sampling. After claim-level
assessment and reweighting, however, accuracy improves by 4.48--7.01 points
over the same unweighted candidates. The gains therefore arise from
falsification-based reliability weighting rather than claim prompting itself,
complementing the matched-call comparison in
Tab.~\ref{tab:main_results}.

\subsection{Rescuing Incorrect Consensus}

\begin{figure}[t]
    \centering
    \includegraphics[width=0.76\textwidth]{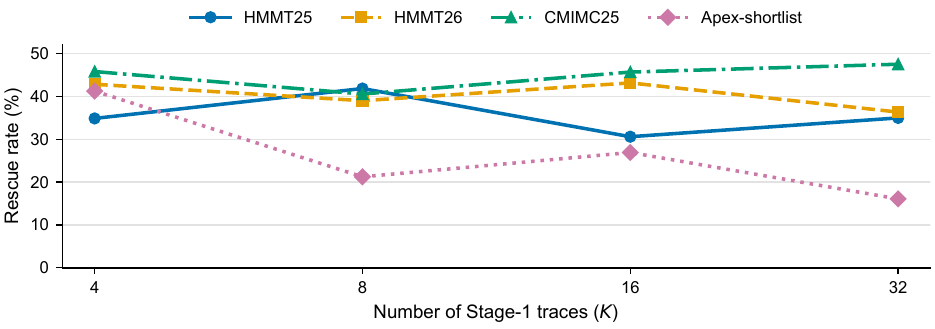}
    \caption{CLR rescue rates on GPT-OSS-20B across Stage-1 sampling counts,
    conditioned on Cons@$K$ being incorrect despite a correct candidate among
    the same $K$ traces.}
    \label{fig:rescue_rate}
\end{figure}

Fig.~\ref{fig:clr_motivation} illustrates CLR overturning an incorrect
count-based consensus when the correct answer is already present among the
candidates. To quantify how often this occurs, we report the rescue rate in
Fig.~\ref{fig:rescue_rate}. For a single TTS flow, the metric is defined as
\begin{equation}
\mathrm{RescueRate@}K =
\frac{
\sum_q\mathbf{1}\!\left[
\exists k:\hat{y}_{q,k}=y_q,\
\hat{y}^{\mathrm{Cons}}_q\neq y_q,\
\hat{y}^{\mathrm{CLR}}_q=y_q
\right]
}{
\sum_q\mathbf{1}\!\left[
\exists k:\hat{y}_{q,k}=y_q,\
\hat{y}^{\mathrm{Cons}}_q\neq y_q
\right]
}.
\label{eq:rescue_rate}
\end{equation}
The denominator counts recoverable consensus failures, in which Cons@$K$ is
incorrect despite the presence of a correct Stage-1 candidate. The numerator
counts the subset corrected by CLR, so $\mathrm{RescueRate@}K$ measures the
fraction of such failures that CLR overturns.

Across the 16 benchmark--budget settings, pooled rescue rates span roughly
16--48\% and average about 37\%, with raw counts in
Appendix~\ref{app:rescue_counts}. Without changing the underlying model or
adding a separate verifier, CLR overturns a substantial fraction of erroneous
consensus outcomes while preserving reliable minority traces.

\subsection{Effect of the Number of Claims}
\begin{table}[H]
    \centering
    \small
    \setlength{\tabcolsep}{4.2pt}
    \begin{tabular}{l cc cc cc}
        \toprule
        \multirow{3}{*}{\textbf{Benchmark}}
        & \multicolumn{6}{c}{\textbf{Number of Claims}} \\
        \cmidrule(lr){2-7}
        & \multicolumn{2}{c}{\textbf{M = 1}}
        & \multicolumn{2}{c}{\textbf{M = 3}}
        & \multicolumn{2}{c}{\textbf{M = 5}} \\
        \cmidrule(lr){2-3}
        \cmidrule(lr){4-5}
        \cmidrule(lr){6-7}
        & Acc.\ (\%) $\uparrow$ & Tok. ($\times 10^3$) $\downarrow$
        & Acc.\ (\%) $\uparrow$ & Tok. ($\times 10^3$) $\downarrow$
        & Acc.\ (\%) $\uparrow$ & Tok. ($\times 10^3$) $\downarrow$ \\
        \midrule
        HMMT25
        & 72.50 & 533.3
        & 75.83 & 623.0
        & \textbf{79.58} & 616.4
        \\
        HMMT26
        & 70.83 & 547.2
        & \textbf{74.62} & 644.2
        & 73.48 & 624.5
        \\
        CMIMC25
        & 77.19 & 511.3
        & 80.62 & 591.7
        & \textbf{82.19} & 583.3
        \\
        Apex-shortlist
        & 20.83 & 682.5
        & 23.96 & 792.6
        & \textbf{24.22} & 795.0 \\
        \bottomrule
    \end{tabular}
    \caption{
        Effect of the number of decision-critical claims on GPT-OSS-20B with
        $K=32$. Accuracy averages eight CLR flows; tokens are mean
        per-problem totals in thousands.
    }
    \label{tab:claim_count_ablation}
\end{table}

In Tab.~\ref{tab:claim_count_ablation}, we examine the claim-set size. At
\(M=1\), Eq.~\ref{eq:trace_reliability} reduces to
\(r_k=v_{k,1}\in\{0,1\}\), yielding binary filtering, while multiple claims
provide broader decision-critical coverage and graded nonlinear weights.
Increasing \(M\) from 1 to 3 improves accuracy by 3.13--3.79 percentage points
across all benchmarks. Increasing \(M\) to 5 yields further gains on three,
while HMMT26 peaks at \(M=3\). The main benefit therefore comes from moving
beyond a single claim, with task-dependent returns thereafter.

The multi-claim settings use 14.1--17.7\% more generated tokens than \(M=1\),
while \(M=3\) and \(M=5\) differ by at most 3.1\% with no consistent direction.
Because \(M\) controls both the number of claims and the exponent in
Eq.~\ref{eq:trace_reliability}, this ablation jointly varies semantic coverage,
score resolution, and penalty sharpness rather than isolating claim count alone.

\begin{figure}[t]
    \centering
    \includegraphics[width=\textwidth]{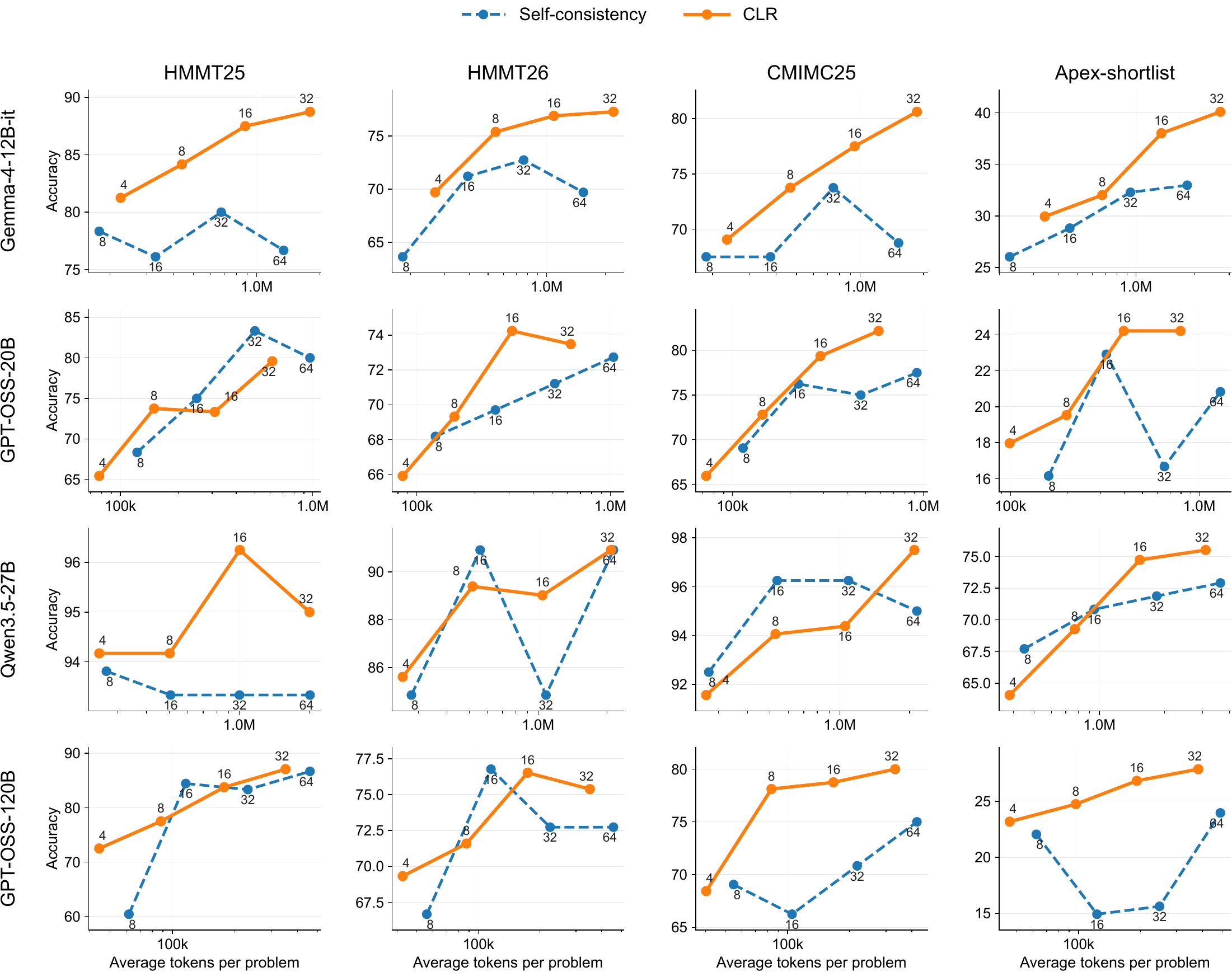}
    \caption{Token--accuracy trade-off across four
    models and four reasoning benchmarks as test-time budgets increase. Compared to count-based self-consistency, CLR avoids early performance saturation and demonstrates a more steady scaling behavior.}
    \label{fig:token_accuracy_frontier}
\end{figure}

\subsection{Cross-Budget Accuracy--Efficiency Scaling}

Fig.~\ref{fig:token_accuracy_frontier} compares CLR with self-consistency
over increasing test-time budgets for all four evaluated models and benchmarks.
The methods allocate requests differently, so the horizontal axis reports
realized token consumption rather than treating equally sized \(K\) values as
equivalent budgets. Unlike the matched-call comparison in
Tab.~\ref{tab:main_results}, these curves show how each method responds as its
test-time budget grows. In several settings, self-consistency saturates or fluctuates as more solutions are added, while CLR improves more steadily, yielding a more favorable accuracy--compute scaling trajectory than count-based aggregation. This
pattern is especially visible across the four Gemma-4-12B-it benchmarks and on
CMIMC25 and Apex-shortlist for both GPT-OSS models, suggesting that the gains are
not confined to the primary operating point.

The curves can nevertheless cross at intermediate budgets, and the two methods
remain closer on the near-saturated Qwen3.5-27B settings; CLR is therefore not
uniformly dominant at every operating point. Across budgets, the value of
additional compute depends on the reliability signal it produces. Claim-level
falsification can move the
accuracy--token frontier outward when additional count-based samples provide
noisy or weakly informative support, while offering less headroom when the base
consensus is already reliable.


\section{Conclusion}

In this work, we explore claim-level falsification as a principle for test-time scaling (TTS). Unlike common forward-search approaches that rely on additional sampling or iterative revision to find a correct reasoning trace, claim-level falsification uses decision-critical semantic anchors to systematically constrain erroneous reasoning paths, allowing reliable paths to exert greater influence. CLR provides an initial instantiation and validation of this idea within consensus-based aggregation framework. Our experiments show that reallocating part of the reasoning budget from additional forward generation to targeted falsification can substantially improve the performance frontier at test time. Future work will explore broader applications of this principle across different TTS paradigms.

\bibliography{references}
\bibliographystyle{colm2026_conference}

\clearpage
\appendix
\section{Experimental Details}
\label{app:experimental_details}

\subsection{Decoding Configuration}
\label{app:decoding_config}

We use the default inference settings of each model unless explicitly stated
otherwise. For all models, we enable thinking mode and keep the thinking effort
at the model default. The same sampling configuration is used for solution
generation and claim-level reliability assessment. Tab.~\ref{tab:decoding_config}
summarizes the model-specific sampling hyperparameters.

\begin{table}[h]
\centering
\small
\begin{tabular}{lcccc}
\toprule
Model & Temperature & Top-\(p\) & Top-\(k\) & Presence penalty \\
\midrule
Gemma-4-12B-it & 1.0 & 1.00 & 40 & 0.0 \\
GPT-OSS-20B & 1.0 & 1.00 & 40 & 0.0 \\
GPT-OSS-120B & 1.0 & 1.00 & 40 & 0.0 \\
Qwen3.5-27B & 1.0 & 0.95 & 20 & 1.5 \\
\bottomrule
\end{tabular}
\caption{Model-specific decoding configurations used in the experiments. All
other model-specific runtime settings are left at their defaults.}
\label{tab:decoding_config}
\end{table}

\subsection{Raw Counts for Rescue Rate}
\label{app:rescue_counts}

Tab.~\ref{tab:rescue_counts} reports the raw counts underlying
Fig.~\ref{fig:rescue_rate}. Because the reported rates pool results from
$N=8$ independent flows, each denominator counts recoverable question--flow
pairs rather than unique benchmark questions. Specifically, it counts pairs
in which at least one Stage-1 candidate has the correct final answer but
unweighted Cons@$K$ is incorrect. The numerator counts the subset corrected by
CLR using the same candidates.

\begin{table}[h]
\centering
\small
\setlength{\tabcolsep}{5pt}
\begin{tabular}{c cccc}
\toprule
$K$ & HMMT25 & HMMT26 & CMIMC25 & Apex-shortlist \\
\midrule
4  & 15/43 (34.88\%) & 15/35 (42.86\%) & 33/72 (45.83\%) & 26/63 (41.27\%) \\
8  & 18/43 (41.86\%) & 16/41 (39.02\%) & 28/69 (40.58\%) & 24/113 (21.24\%) \\
16 & 15/49 (30.61\%) & 19/44 (43.18\%) & 32/70 (45.71\%) & 42/156 (26.92\%) \\
32 & 14/40 (35.00\%) & 20/55 (36.36\%) & 29/61 (47.54\%) & 32/199 (16.08\%) \\
\bottomrule
\end{tabular}
\caption{Raw counts for the CLR rescue rate. Each entry reports rescued cases over
recoverable consensus errors, followed by the rescue rate in parentheses.}
\label{tab:rescue_counts}
\end{table}

\section{Prompt Templates and Output Formats}
\label{app:prompts}

This appendix reports the base prompt used for regular solution sampling and
the two prompt templates used by CLR. The placeholder
\verb|{question}| is replaced by the benchmark problem, and
\verb|{num_claims}| corresponds to \(M\) in the main text and specifies the
number of claims and verdicts. The XML tag sequences and
\verb|{claims_block}| are assembled dynamically according to this
hyperparameter. Both stages use the same sampling parameters.

\subsection{Regular Solution Sampling}
\label{app:regular_prompt}

Regular sampling uses the following base prompt. CLR Stage 1 extends this
template by appending the claim-generation instructions shown in the next
subsection.

\begin{Verbatim}[
    breaklines=true,
    breakanywhere=true,
    fontsize=\scriptsize,
    frame=single,
    framesep=2mm,
    xleftmargin=2mm,
    xrightmargin=2mm
]
{question}

Let's think step by step and output the final answer within \boxed{}.
\end{Verbatim}

\subsection{Stage 1: Solution and Claim Generation}
\label{app:stage1_prompt}

Stage 1 asks the model to generate a step-by-step solution, place the final
answer in \verb|\boxed{}|, and append a fixed number of verification claims.
The template, including its runtime placeholders, is shown below.

\begin{Verbatim}[
    breaklines=true,
    breakanywhere=true,
    fontsize=\scriptsize,
    frame=single,
    framesep=2mm,
    xleftmargin=2mm,
    xrightmargin=2mm
]
{question}

Let's think step by step and output the final answer within \boxed{}.

After the final answer, append exactly **{num_claims}** verification claims selected from your completed 
solution.

### Claim Rules
- **Relevance:** Each claim should be an intermediate mathematical statement used to reach the final answer.
- **Diagnostic Value:** Prefer claims whose failure would strongly undermine the final answer.
- **Non-triviality:** Do not use claims that merely restate the problem, repeat the final answer, duplicate 
another claim, or only describe a trivial final arithmetic step.
- **Clarity:** Use precise mathematical objects, variables, equations, constraints, transformations, case 
distinctions, or counting formulas.

The final answer must appear before the claims and must be written as:
\boxed{final answer}

Then output exactly:
<claims>
{claim_tags}
</claims>
\end{Verbatim}

For compactness, \verb|{claim_tags}| denotes exactly
\verb|{num_claims}| entries of the form
\texttt{<claim>CLAIM\_CONTENT</claim>}, assembled at runtime.


\subsection{Stage 2: Claim-Level Reliability Assessment}
\label{app:stage2_prompt}

Stage 2 stress-tests the extracted claims and returns an ordered binary
verdict for each claim. The placeholder \verb|{claims_block}| contains the
claims extracted in Stage 1. The template is shown below.

\begin{Verbatim}[
    breaklines=true,
    breakanywhere=true,
    fontsize=\scriptsize,
    frame=single,
    framesep=2mm,
    xleftmargin=2mm,
    xrightmargin=2mm
]
You are a rigorous mathematical verifier.

## Problem
{question}

## Claims to Verify
The following {num_claims} claims were made during a solution attempt:
{claims_block}

## Your Task
Treat every claim as a suspicious assertion rather than a fact.

For each claim, try to FALSIFY it by checking whether it:
- contradicts the problem conditions,
- conflicts with another claim in the list,
- relies on an unjustified inference or hidden assumption,
- fails on a simple test case or counterexample,
- contains a calculation or logical error.

Do NOT assume that a claim is true just because you cannot immediately find a counterexample. Actively 
search for flaws.

After your analysis of each claim, give a final verdict.

## Output Format
After your analysis, you MUST end your response with verdicts in this format:

<verdicts>
{verdict_tags}
</verdicts>

Output exactly {num_claims} verdict tags in order. Replace each VERDICT with exactly one token: VALID or 
REFUTED.
Do not output VERDICT literally. Do not output "VALID or REFUTED" literally.
Do not put any text after </verdicts>.
\end{Verbatim}

For compactness, \verb|{verdict_tags}| denotes exactly
\verb|{num_claims}| entries of the form
\texttt{<verdict>VERDICT</verdict>}, assembled at runtime.

\paragraph{Claim rendering.}
Before insertion into the prompt, line breaks within each extracted claim are
collapsed to spaces. The dynamically assembled \verb|{claims_block}| has the
following form:
\begin{Verbatim}[
    fontsize=\scriptsize,
    frame=single,
    framesep=2mm,
    xleftmargin=2mm,
    xrightmargin=2mm
]
1. <claim text 1>
2. <claim text 2>
...
{num_claims}. <claim text {num_claims}>
\end{Verbatim}

\paragraph{Output parsing and fallbacks.}
Malformed XML tags or truncated traces may prevent complete claim extraction.
If fewer than \(M\) claims are successfully parsed, the remaining slots are
padded with empty claims, while missing or unparseable verdicts default to not
refuted.

\end{document}